\documentclass[12pt]{article}

\usepackage{sbc-template}
\usepackage{graphicx,url}
\usepackage{subcaption}
\usepackage[utf8]{inputenc}
\usepackage[english]{babel}
\usepackage{amsmath}
\usepackage{multirow}
\usepackage{booktabs} 
\usepackage{colortbl}
\usepackage{arydshln}
\usepackage{xcolor}
\usepackage{amsfonts}
\title{Does Machine Unlearning Preserve Clinical Safety?\\ A Risk Analysis for Medical Image Classification}

\author{Andreza M. C. Falcao\inst{1}, Filipe R. Cordeiro\inst{1}}

\address{
  Visual Computing Lab, Departamento de Computação, \\ Universidade Federal Rural de Pernambuco (UFRPE), Brasil
  \email{andreza.mcfalcao@ufrpe.br, filipe.rolim@ufrpe.br } 
}

\begin{document} 

\maketitle

\begin{abstract}

The application of Deep Learning in medical diagnosis must balance patient safety with compliance with data protection regulations. Machine Unlearning enables the selective removal of training data from deployed models. However, most methods are validated primarily through efficiency and privacy-oriented metrics, with limited attention to clinically asymmetric error costs. In this work, we investigate how unlearning affects clinical risk in binary medical image classification. We show that standard unlearning strategies (Fine-Tuning, Random Labeling, and SalUn) may reduce test utility while increasing false-negative rates, thereby amplifying clinical risk. To mitigate this, we propose SalUn-CRA (Clinical Risk-Aware), a variant of SalUn that replaces random relabeling with entropy-based forgetting for malignant samples in the forget set, preventing the model from learning harmful benign associations. We evaluate on DermaMNIST and PathMNIST medical image datasets under 20\% and 50\% data removal. Using Global Risk metrics with asymmetric costs, SalUn-CRA achieves lower or comparable clinical risk to full retraining while preserving unlearning effectiveness. These results suggest that clinical risk should be an integral component of unlearning validation in medical systems.

\end{abstract}
     
\section{Introduction}

The integration of Deep Learning into clinical workflows has significantly improved diagnostic performance in medical image analysis, achieving results comparable to human specialists in tasks such as skin lesion and histopathological classification~\cite{Chan2020}. However, these models typically rely on large volumes of training data containing sensitive patient information, making them subject to strict data protection regulations.

Recent regulatory frameworks, including the Brazilian General Data Protection Law (LGPD)~\cite{lgpd} and the European General Data Protection Regulation (GDPR)~\cite{gdpr}, establish the \textit{Right to be Forgotten}~\cite{Hoofnagle2019}, requiring the removal of personal data upon request. In the context of Machine Learning, removing samples from the dataset alone is insufficient if the model has already been trained on them, as the influence of those samples must also be removed from the model parameters. Full retraining on the remaining data is the most straightforward solution, but it is computationally prohibitive for large-scale datasets or frequent removal requests~\cite{mester2024medical}. Machine Unlearning (MU) addresses this challenge by developing algorithms that update model weights to forget specific data without full retraining~\cite{fan2024salun}.

Although substantial progress has been made in MU methods~\cite{warnecke2021, golatkar2020eternal, fan2024salun}, their evaluation has primarily focused on computational efficiency, privacy guarantees, and similarity to retrained models~\cite{zhang2023review}. MU approaches applied to medical domains have also been restricted to accuracy-based evaluation~\cite{sakib2024unlearning, maverick2025}. These protocols implicitly assume symmetric error costs, but this assumption does not hold in medical diagnosis~\cite{ling2010cost}. A False Negative (failure to detect a malignant condition) may delay treatment and impact patient survival, whereas a False Positive typically results in additional confirmatory tests~\cite{scholz2024imbalance}. Relying solely on global metrics may obscure clinically critical degradations in sensitivity after unlearning. Importantly, it remains unclear whether MU procedures preserve, amplify, or mitigate clinical risk.

Despite the growing literature on Machine Unlearning, a critical question remains unanswered: \textbf{Do state-of-the-art machine unlearning methods preserve clinical safety?} To the best of our knowledge, no prior work has systematically investigated how unlearning affects cost-sensitive clinical risk in medical image classification.


In this work, we evaluate machine unlearning methods under clinical risk formulations that model asymmetric diagnostic costs. We propose \textit{SalUn-CRA} (Clinical Risk-Aware), a modification of SalUn~\cite{fan2024salun} that prevents harmful benign reassignment of malignant samples during forgetting, and introduce Global Risk metrics based on the asymmetric costs of false positives and false negatives. Our main contributions are:

\begin{itemize}
    \item We introduce \textit{Global Risk} as an evaluation criterion for medical unlearning, explicitly modelling asymmetric clinical costs.
    \item We show that common unlearning strategies (Fine-Tuning, Random Labeling, and SalUn) may increase clinical risk by trading recall for specificity, leading to higher false negative rates.
    \item We propose SalUn-CRA, a risk-aware modification of SalUn that applies entropy-based forgetting to malignant forget samples instead of random relabeling, mitigating risk amplification while preserving unlearning behaviour.
\end{itemize}

We evaluate our approach on two binary classification tasks derived from MedMNIST~\cite{medmnist}: DermaMNIST (skin lesion classification) and PathMNIST (colorectal tissue classification), under 20\% and 50\% data removal. 
We focus on binary classification (malignant vs.\ benign) because it represents the most critical decision point in clinical pipelines, carrying the highest asymmetry in misclassification costs~\cite{ling2010cost}, and enables a clean risk analysis with only two cost parameters, avoiding the combinatorial complexity of multi-class cost matrices.

\section{Prior Work}

\subsection{Machine Unlearning: Foundations and Methods}

Machine Unlearning approaches can be categorized into exact and approximate methods. Exact methods ensure the complete removal of a specific data subset and aim to retrain the model at lower computational cost than full retraining~\cite{li2024taxonomy}. Techniques within this class typically leverage checkpoint-based retraining strategies, selectively updating portions of the model rather than performing full retraining~\cite{bourtoule2021machine}. Although more computationally efficient than complete retraining, exact methods remain prohibitively costly for frequent or large-scale unlearning tasks.


Among the foundational approaches, \textit{Amnesiac Learning}~\cite{graves2021amnesiac} stores per-batch gradient contributions during training and reverses them during unlearning, selectively eliminating the influence of the target data. Golatkar et al.~\cite{golatkar2020eternal} leverage Fisher information to estimate which model weights encode information about the samples to be removed and modify them directly. Barez et al.~\cite{barez2025open} argue that unlearning can degrade existing safety mechanisms and produce unintended side effects on model behaviour. This observation directly motivates this work's concern with clinical risk after data removal.
Fine-Tuning (FT)~\cite{warnecke2021} continues training on the retain set only, allowing the model to gradually forget removed samples through parameter drift. Random Labeling (RL)~\cite{golatkar2020eternal} assigns random labels to forget samples, forcing the model to unlearn their correct associations. Both methods are computationally efficient but lack mechanisms to preserve class-specific performance.

Saliency Unlearning (SalUn)~\cite{fan2024salun} represents one of the current state-of-the-art (SOTA) approaches in approximate unlearning. It computes a saliency map based on the gradients of the forgetting loss to identify the weights most relevant to the target data. Then it applies random labelling only to these salient parameters. 

\subsection{Machine Unlearning in Healthcare}


The application of Deep Learning to medical image analysis is well established~\cite{chan2020deep}, yet its intersection with machine unlearning remains underexplored. Nasirigerdeh et al.~\cite{nasirigerdeh2024machine} studied unlearning in medical images, focusing on re-identification attacks, but without addressing diagnostic metrics. Hardan et al.~\cite{hardan2025forget} proposed Forget-MI for multimodal medical data. Falc\~ao and Cordeiro~\cite{cordeiro25} evaluated MU on medical datasets, though limited to standard MU metrics.


Despite the growing research on machine unlearning for healthcare applications, a critical gap persists between the evaluation practices of the unlearning community and the requirements of clinical machine learning. None of the existing healthcare-oriented unlearning studies reports sensitivity and specificity as separate metrics, nor do they employ Balanced Accuracy to account for class imbalance commonly present in medical datasets. More importantly, no existing work evaluates the clinical risk implications of the unlearning process, that is, whether unlearning disproportionately degrades the model's ability to detect positive (pathological) cases, thereby increasing false negative rates in a clinically dangerous manner. This paper aims to bridge this gap by explicitly evaluating unlearning methods under clinical risk metrics. While previous works focus on privacy and computational efficiency, we investigate whether data removal compromises diagnostic safety.



\section{Methodology}

\subsection{Problem Formulation}

Let $\mathcal{D}=\{x_i,y_i\}_{i=1}^N$ be a training dataset of images $x_i$ with associated labels $y_i \in \{1,\dots,K\}$, where $K$ is the number of classes and $N$ the total number of samples. The Machine Unlearning problem can be formally defined as the task of removing the influence of a specific subset of data $\mathcal{D}_f \subset \mathcal{D}$ from a previously trained model $\theta_o = \mathcal{A}(\mathcal{D})$, where $\theta_o$ is the set of weights resulting from applying a training algorithm $\mathcal{A}$ to the dataset $\mathcal{D}$. Traditional retraining approaches retrain the model on the remaining subset $\mathcal{D}_r = \mathcal{D} \setminus \mathcal{D}_f$, obtaining the retrained model weights $\theta_r = \mathcal{A}(\mathcal{D}_r)$, retrained from scratch without the data subset $\mathcal{D}_f$.

Given this formulation, the MU task consists of employing an unlearning algorithm $\mathcal{U}$, which, starting from the trained model $\theta_o$, the subset to be forgotten $\mathcal{D}_f$, and the remaining subset $\mathcal{D}_r$, produces a new model $\theta_u = \mathcal{U}(\theta_o, \mathcal{D}_f, \mathcal{D}_r)$. It is expected that $\theta_u$ approximates, in terms of output distribution, the ideal retrained model $\theta_r$.

The performance of the unlearned model is evaluated through the metric gap, defined as $MG = |\mathcal{M}_{MU} - \mathcal{M}_{retrain}|$, where $\mathcal{M}_{MU}$ represents the evaluation metric on the unlearned model and $\mathcal{M}_{retrain}$ corresponds to the metric on the retrained model. The main challenge lies in designing unlearning algorithms that are not only effective at removing the influence of $\mathcal{D}_f$ but also significantly more computationally efficient than full retraining while preserving performance on the remaining dataset $\mathcal{D}_r$.

\subsection{SalUn-CRA: Clinical Risk-Aware Saliency Unlearning}

Saliency Unlearning (SalUn)~\cite{fan2024salun} is one of the current state-of-the-art methods in approximate machine unlearning. Its core contribution is the introduction of \emph{weight saliency} into the unlearning process. Rather than modifying all model parameters during unlearning, SalUn identifies the subset of weights most relevant to the forget set $\mathcal{D}_f$ through a gradient-based saliency map:
\begin{equation}
    \mathbf{m}_S = \mathbb{1}\!\left(\left|\nabla_{\boldsymbol{\theta}} \ell_f(\boldsymbol{\theta}; \mathcal{D}_f)\big|_{\boldsymbol{\theta}=\boldsymbol{\theta}_o}\right| \geq \gamma\right),
    \label{eq:saliency_mask}
\end{equation}
where $\ell_f$ is the forgetting loss (cross-entropy on $\mathcal{D}_f$), $\boldsymbol{\theta}_o$ denotes the original model weights, $\gamma$ is a hard threshold (set to the median of the gradient magnitudes), and $\mathbb{1}(\cdot)$ is an element-wise indicator function. This mask decomposes the model into \emph{salient weights}, which are updated during unlearning, and \emph{intact weights}, which remain unchanged:
\begin{equation}
    \boldsymbol{\theta}_u = \mathbf{m}_S \odot (\Delta\boldsymbol{\theta} + \boldsymbol{\theta}_o) + (\mathbf{1} - \mathbf{m}_S) \odot \boldsymbol{\theta}_o,
    \label{eq:salun_decomposition}
\end{equation}
where $\odot$ denotes the element-wise product. Once the salient weights are identified, SalUn applies Random Labeling (RL) exclusively to these parameters. For each sample $(x_i, y_i) \in \mathcal{D}_f$, a random label $y_i' \neq y_i$ is assigned, and the model is fine-tuned on the relabeled forget set combined with a regularization term on the retain set $\mathcal{D}_r$:
\begin{equation}
    \mathcal{L}_{\text{SalUn}}(\boldsymbol{\theta}_u) = \mathbb{E}_{(x,y)\sim\mathcal{D}_f, y'\neq y}\left[\ell_{\text{CE}}(\boldsymbol{\theta}_u; x, y')\right] + \alpha\,\mathbb{E}_{(x,y)\sim\mathcal{D}_r}\left[\ell_{\text{CE}}(\boldsymbol{\theta}_u; x, y)\right],
    \label{eq:salun_loss}
\end{equation}
where $\alpha > 0$ balances unlearning effectiveness and model utility preservation.

While SalUn achieves strong unlearning performance on general-purpose benchmarks~\cite{fan2024salun}, its reliance on uniform random relabeling introduces a critical problem in binary medical classification settings where class~1 corresponds to the malignant (positive) class and class~0 to the benign (negative) class.


While SalUn achieves strong performance on general-purpose benchmarks~\cite{fan2024salun}, its reliance on random relabeling introduces a critical problem in binary medical classification. In the binary case, every malignant sample in $\mathcal{D}_f$ deterministically receives the label ``benign'', unlike multi-class scenarios where errors distribute across $C-1$ classes. The model is thus actively trained to associate malignant features with the benign class.

This relabeling contamination has a direct and asymmetric impact on the decision boundary. The model learns to suppress activations for malignant patterns, shifting the classification threshold toward higher specificity at the expense of sensitivity. In clinical terms, this translates to an increase in false negatives: malignant lesions that the model now classifies as benign. While the overall accuracy or balanced accuracy may appear stable,
missed malignant diagnoses carry far greater consequences than false alarms~\cite{ling2010cost}.

To mitigate this risk, we propose SalUn-CRA (Clinical Risk-Aware), a modification of SalUn that applies class-dependent forgetting strategies within the forget set. The key idea is to decouple the forgetting mechanism based on the clinical severity of each class. For malignant samples ($y_i = 1$) in $\mathcal{D}_f$, instead of random relabeling, we use a \emph{maximum entropy loss} that pushes the model's output distribution toward uniformity, encouraging maximum uncertainty without assigning a benign label. For benign samples ($y_i = 0$) in $\mathcal{D}_f$, standard random relabeling is applied as in the original SalUn. In the binary case, benign samples receive the malignant label, which does not introduce the same clinical risk.

Let $\mathcal{D}_f^{+} = \{(x_i, y_i) \in \mathcal{D}_f \mid y_i = 1\}$ and $\mathcal{D}_f^{-} = \{(x_i, y_i) \in \mathcal{D}_f \mid y_i = 0\}$ denote the malignant and benign subsets of the forget set, respectively. The maximum entropy loss for malignant samples is defined as:
\begin{equation}
    \mathcal{L}_{\text{entropy}}(\boldsymbol{\theta}; x) = -\sum_{c=1}^{C} p_c(x;\boldsymbol{\theta}) \log p_c(x;\boldsymbol{\theta}),
    \label{eq:entropy_loss}
\end{equation}
where $p_c(x;\boldsymbol{\theta}) = \text{softmax}(f_{\boldsymbol{\theta}}(x))_c$ is the predicted probability for class $c$, and $C$ is the number of classes. Maximizing this entropy drives the output toward a uniform distribution $p_c = 1/C$ for all $c$, achieving maximum prediction uncertainty.

The complete SalUn-CRA loss function combines three components:
\begin{equation}
    \mathcal{L}_{\text{CRA}}(\boldsymbol{\theta}_u) = 
    -\underbrace{\mathbb{E}_{x \in \mathcal{D}_f^{+}}\!\left[\mathcal{L}_{\text{entropy}}(\boldsymbol{\theta}_u; x)\right]}_{\text{entropy maximization (malignant)}}
    + \underbrace{\mathbb{E}_{(x,y') \in \mathcal{D}_f^{-}}\!\left[\ell_{\text{CE}}(\boldsymbol{\theta}_u; x, y')\right]}_{\text{random labeling (benign)}}
    + \underbrace{\alpha\,\mathbb{E}_{(x,y) \in \mathcal{D}_r}\!\left[\ell_{\text{CE}}^{w}(\boldsymbol{\theta}_u; x, y)\right]}_{\text{retain regularization}},
    \label{eq:cra_loss}
\end{equation}
where $y'$ is the random label assigned to benign forget samples ($y' = 1$ in the binary case), $\ell_{\text{CE}}^{w}$ denotes weighted cross-entropy with inverse-frequency class weights to address class imbalance, and $\alpha > 0$ controls the retain regularization strength. The weight saliency mask $\mathbf{m}_S$ from Equation~\eqref{eq:saliency_mask} is applied identically as in the original SalUn, restricting parameter updates to the salient subset.

\subsection{Datasets}
\label{sec:datasets}

We use DermaMNIST and PathMNIST, two datasets from the MedMNIST collection~\cite{medmnist}, converted to binary classification tasks. All images were resized to $128\times128$ pixels to standardize input across experiments.

DermaMNIST consists of 10,015 dermatoscopic images of pigmented skin lesions from the HAM10000 dataset. The original dataset contains 7 diagnostic categories. We binarize the labels by grouping melanoma, basal cell carcinoma, and actinic keratosis as \textit{malignant}, while melanocytic nevus, benign keratosis, vascular lesion, and dermatofibroma are grouped as \textit{benign}. 

PathMNIST contains 107,180 histopathological images of colorectal tissue. The original 9-class problem was binarized by grouping cancer-associated stroma and colorectal adenocarcinoma as malignant, and all other tissue types (adipose, background, debris, lymphocytes, mucus, smooth muscle, and normal colon mucosa) as benign.


We adopt this binarization protocal because it serves a methodological purpose in the context of this study. By reducing the problem to two classes, we isolate the effect of unlearning on the sensitivity--specificity trade-off, which is the core mechanism through which clinical risk is amplified.
The binary formulation provides a controlled experimental framework where shifts in false negative and false positive rates can be directly measured and linked to the forgetting procedure. Table~\ref{tab:dataset_stats} summarizes the binarization protocol for both datasets.


\begin{table}[h]
\centering
\caption{Dataset statistics of DermaMNIST and PathMNIST after binarization. Original multi-class labels are grouped into clinically meaningful binary categories.}
\label{tab:dataset_stats}
\scalebox{0.8}{
\begin{tabular}{l|cc|cc}
\hline
\multirow{2}{*}{Split} & \multicolumn{2}{c|}{DermaMNIST} & \multicolumn{2}{c}{PathMNIST} \\
 & Benign & Malignant & Benign & Malignant \\
\hline
Train & 5,641 (80.5\%) & 1,366 (19.5\%) & 67,710 (75.2\%) & 22,286 (24.8\%) \\
Val & 807 (80.5\%) & 196 (19.5\%) & 7,527 (75.2\%) & 2,477 (24.8\%) \\
Test & 1,613 (80.4\%) & 392 (19.6\%) & 5,526 (77.0\%) & 1,654 (23.0\%) \\
\hline
\end{tabular}
}
\end{table}



\subsection{Unlearning Scenarios}

The goal of unlearning is to update a model trained on $\mathcal{D}_{train}$ so that it behaves as if trained only on $\mathcal{D}_{retain} = \mathcal{D}_{train} \setminus \mathcal{D}_{forget}$. We evaluate two removal scenarios: 20\% and 50\% of training sample removal, following the protocol in~\cite{fan2024salun, cordeiro25}.
In both scenarios, we apply \textit{balanced removal}, where samples are removed proportionally from each class, preserving the original class distribution. This isolates the effect of data reduction from artificial distribution shift.

\subsection{Unlearning Methods}

We evaluate the following unlearning methods: Retrain, Fine-Tuning (FT)~\cite{warnecke2021}, Random Labeling (RL), Saliency Unlearning (Salun) and the proposed Salun-CRA. 

Retrain approch is the complete retraining from scratch on $\mathcal{D}_{retain}$, serving as the reference for ideal unlearning behavior. Fine-Tuning continues training the original model on $\mathcal{D}_{retain}$ only, allowing gradual forgetting through parameter updates.
Random Labeling (RL) assigns random labels to samples in $\mathcal{D}_{forget}$ and trains on the combined dataset, forcing the model to unlearn correct associations. SalUn computes a saliency mask identifying parameters most relevant to $\mathcal{D}_{forget}$, then applies random labeling only to these salient weights.
SalUn-CRA is our clinical risk-aware modification of SalUn, as described in Section~3.2. 

\subsection{Evaluation Metrics}
\label{sec:metrics}

We evaluate unlearning methods across three dimensions: model utility, unlearning effectiveness, and clinical risk.
For model utility, we use Especificity, Recall,  Balanced Accuracy (BAC) and Area Under the ROC Curve (AUC). Specificity measures the proportion of benign cases correctly identified by $TN / (TN + FP)$, where $TN$ and $FP$ represent the number true negatives and false positives, respectivelly. Recall measures the proportion of malignant cases correctly identified by the equation $TP / (TP + FN)$, with $FN$ being the number of false negatives. BAC IS defined as the average of specificity and recall, which is robust to class imbalance. AUC is the area under the ROC curve, measuring overall discrimination ability.

The standard evaluation protocol in MU is based on accuracy values computed on the retain  and forget sets~\cite{fan2024salun}. However, we adapted these metrics to use BAC instead of Accuracy to better reflect performance on imbalanced medical datasets. The adapted metrics are: UBAC, the balanced accuracy on $\mathcal{D}_{forget}$ (lower values indicate successful forgetting); RBAC, the balanced accuracy on $\mathcal{D}_{retain}$ (higher values indicate preserved utility); TBAC, the balanced accuracy on the test set; MIA, the Membership Inference Attack accuracy (lower values indicate better privacy); and GAP, the average absolute difference between unlearned and retrained model metrics (lower GAP indicates closer approximation to the gold standard).

We propose a clinical risk formulation based on cost-sensitive evaluation~\cite{haimerl2025}, capturing the asymmetric cost structure of medical diagnosis. The general Global Risk is defined as:

\begin{equation}
\text{Risk} = \frac{C_{\mathit{FP}} \cdot \mathit{FP} + C_{\mathit{FN}} \cdot \mathit{FN}}{N}
\end{equation}
where $\mathit{FP}$ and $\mathit{FN}$ are the number of false positives and false negatives, respectively, $N$ is the total number of samples, and $C_{\mathit{FP}}$ and $C_{\mathit{FN}}$ are the misclassification costs for each error type.
For this purpose, we consider two risk scenarios: \textbf{Global Risk I} and \textbf{Global Risk II}.  Global Risk I sets $C_{FN} = 1$ and $C_{FP} = 1$, representing equal misclassification costs, as commonly assumed in the  literature. Global Risk II sets $C_{FN} = 20$ and $C_{FP} = 1$, reflecting a clinically realistic scenario where missed malignant diagnoses carry substantially higher costs than false positives. The adoption of $C_{FN} = 20$ for Risk~II is a representative value for serious disease screening scenarios, as there is no universally agreed-upon cost ratio in the literature, as it depends on context.
Lower risk values indicate safer models. Global Risk II specifically penalizes methods that sacrifice sensitivity (recall) for specificity, as this trade-off increases the number of missed malignancies.

\subsection{Implementation}

We adopt ResNet-18~\cite{wu2019wider} as the backbone architecture for all experiments, following standard practice in medical imaging tasks~\cite{cordeiro25}. The baseline model is trained for 200 epochs using SGD optimizer with learning rate 0.1, momentum 0.9, and batch size 256.

To address class imbalance and incorporate the clinical priority of detecting malignant cases, all models are trained with Weighted Cross-Entropy Loss:
\begin{equation}
\mathcal{L}_w = -\frac{1}{N} \sum_{i=1}^{N} w_{y_i} \log(p(y_i | x_i))
\end{equation}
where $w_{y_i}$ is the weight assigned to the true class $y_i$. Weights are defined inversely proportional to class frequencies in the training set, ensuring that the minority class has greater influence on gradient updates. 

All unlearning methods are executed for 10 epochs with learning rate 0.01, following the protocol in~\cite{fan2024salun}.

\section{Results}
\label{sec:results}

 We analyze the trade-off between unlearning effectiveness, model utility, and clinical risk across DermaMNIST and PathMNIST under 20\% and 50\% data removal scenarios. We evaluated the SOTA MU models Fine-Tuning (FT)~\cite{warnecke2021}, Random Labeling (RL)~\cite{golatkar2020eternal}, Salun~\cite{fan2024salun} and the proposed Salun-CRA using the metrics described in section~\ref{sec:metrics}. 

\subsection{Utility and Unlearning Metrics}

Table~\ref{tab:resultados} presents comprehensive results for utility and unlearning metrics. Regarding model utility, SalUn-CRA achieves the highest Balanced Accuracy (BAC) in three out of four scenarios. On DermaMNIST with 50\% removal, SalUn-CRA ties with Fine-Tuning at 0.81 BAC.

\begin{table*}[htbp]
\centering
\caption{Machine unlearning results on binary DermaMNIST and PathMNIST datasets. 
Best results among approximate methods highlighted with gray background. 
Values in parentheses represent the difference relative to Retrain.}
\label{tab:resultados}
\scalebox{0.75}{
\begin{tabular}{l cccc cccc}
\toprule
 & \multicolumn{4}{c}{\textbf{Utility Metrics}} & \multicolumn{4}{c}{\textbf{Unlearning Metrics}} \\
 \cmidrule(lr){2-5} \cmidrule(lr){6-9}
 \textbf{Method} & Spec. $\uparrow$ & Recall $\uparrow$ & BAC $\uparrow$ & AUC $\uparrow$ & UBAC $\downarrow$ & RBAC $\uparrow$ & TBAC $\uparrow$ & MIA $\downarrow$\\
\midrule
\multicolumn{9}{l}{\textit{DermaMNIST -- 20\% removal}} \\
\midrule
 Retrain & 0.83 & 0.77 & 0.80 & 0.90 & 0.24 (0.00) & 0.92 (0.00) & 0.80 (0.00) & 21.27 (0.00) \\
 \hdashline
 FT & 0.87 & 0.71 & 0.79 & 0.90 & 0.10 (0.14) & \cellcolor{gray!25}\textbf{0.96} (0.04) & 0.79 (0.01) & 10.78 (10.49) \\
 RL & \cellcolor{gray!25}\textbf{0.90} & 0.68 & 0.79 & \cellcolor{gray!25}\textbf{0.90} & \cellcolor{gray!25}\textbf{0.04} (0.20) & 0.95 (0.03) & 0.79 (0.01) & \cellcolor{gray!25}\textbf{5.50} (15.77) \\
 SalUn & 0.88 & 0.72 & 0.80 & \cellcolor{gray!25}\textbf{0.90} & \cellcolor{gray!25}\textbf{0.04} (0.02) & 0.95 (0.03) & 0.80 (0.00) & 7.21 (14.06)\\
 SalUn-CRA (Ours) & 0.85 & \cellcolor{gray!25}\textbf{0.78} & \cellcolor{gray!25}\textbf{0.81} & 0.90 & 0.08 (0.16) & 0.94 (0.02) & \cellcolor{gray!25}\textbf{0.81} (0.01) & 17.77 (3.50) \\
\midrule
\multicolumn{9}{l}{\textit{DermaMNIST -- 50\% removal}} \\
\midrule
 Retrain & 0.81 & 0.76 & 0.79 & 0.87 & 0.22 (0.00) & 0.89 (0.00) & 0.79 (0.00) & 21.61 (0.00) \\
 \hdashline
 FT & 0.86 & 0.75 & \cellcolor{gray!25}\textbf{0.81} & 0.90 & 0.09 (0.13) & \cellcolor{gray!25}\textbf{0.96} (0.07) & 0.80 (0.01) & 17.53 (4.08) \\
 RL & \cellcolor{gray!25}\textbf{0.87} & 0.74 & 0.80 & 0.90 & \cellcolor{gray!25}\textbf{0.05} (0.17) & 0.95 (0.06) & 0.80 (0.01) & \cellcolor{gray!25}\textbf{10.13} (11.48) \\
 SalUn & 0.86 & 0.75 & 0.80 & \cellcolor{gray!25}\textbf{0.91} & 0.06 (0.16) & 0.95 (0.06) & 0.80 (0.01) & 13.16 (8.45) \\
 SalUn-CRA (Ours) & 0.83 & \cellcolor{gray!25}\textbf{0.79} & \cellcolor{gray!25}\textbf{0.81} & 0.90 & 0.08 (0.14) & 0.94 (0.05) & \cellcolor{gray!25}\textbf{0.81} (0.02) & 18.73 (2.88) \\
\midrule
\multicolumn{9}{l}{\textit{PathMNIST -- 20\% removal}} \\
\midrule
 Retrain & 0.99 & 0.91 & 0.95 & 0.99 & 0.00 (0.00) & 1.00 (0.00) & 0.95 (0.00) & 1.64 (0.00) \\
 \hdashline
 FT & \cellcolor{gray!25}\textbf{0.99} & 0.89 & 0.94 & 0.98 & \cellcolor{gray!25}\textbf{0.00} (0.00) & 1.00 (0.00) & 0.94 (0.01) & \cellcolor{gray!25}\textbf{1.21} (0.43)\\
 RL & 0.98 & 0.91 & 0.95 & 0.99 & \cellcolor{gray!25}\textbf{0.00} (0.00) & 1.00 (0.00) & 0.95 (0.00) & 1.65 (0.01) \\
 SalUn & 0.98 & 0.91 & 0.94 & 0.98 & 0.01 (0.01) & 1.00 (0.00) & 0.94 (0.01) & 1.52 (0.12)\\
 SalUn-CRA (Ours) & 0.98 & \cellcolor{gray!25}\textbf{0.92} & \cellcolor{gray!25}\textbf{0.95} & \cellcolor{gray!25}\textbf{0.99} & \cellcolor{gray!25}\textbf{0.00} (0.00) & 1.00 (0.00) & \cellcolor{gray!25}\textbf{0.95} (0.00) & 1.49 (0.15) \\
\midrule
\multicolumn{9}{l}{\textit{PathMNIST -- 50\% removal}} \\
\midrule
 Retrain & 0.97 & 0.90 & 0.93 & 0.98 & 0.01 (0.00) & 1.00 (0.00) & 0.93 (0.00) & 2.06 (0.00) \\
 \hdashline
 FT& \cellcolor{gray!25}\textbf{0.99} & 0.89 & \cellcolor{gray!25}\textbf{0.94} & 0.98 & 0.00 (0.01) & 1.00 (0.00) & \cellcolor{gray!25}\textbf{0.94} (0.01) & 1.53 (0.53) \\
 RL & \cellcolor{gray!25}\textbf{0.99} & 0.89 & \cellcolor{gray!25}\textbf{0.94} & \cellcolor{gray!25}\textbf{0.98} & 0.00 (0.01) & 1.00 (0.00) & \cellcolor{gray!25}\textbf{0.94} (0.01) & 1.34 (0.72) \\
 SalUn & \cellcolor{gray!25}\textbf{0.99} & 0.88 & 0.93 & 0.98 & 0.00 (0.01) & 1.00 (0.00) & 0.93 (0.00) & \cellcolor{gray!25}\textbf{1.19} (0.87) \\
 SalUn-CRA (Ours) & 0.98 & \cellcolor{gray!25}\textbf{0.91} & \cellcolor{gray!25}\textbf{0.94} & \cellcolor{gray!25}\textbf{0.98} & 0.00 (0.01) & 1.00 (0.00) & \cellcolor{gray!25}\textbf{0.94} (0.01) & 1.98 (0.08) \\
\bottomrule
\end{tabular}
}
\end{table*}

More importantly, SalUn-CRA consistently achieves the highest Recall across all scenarios. This preservation of recall to malignant cases is important to reduce te clinical risk, wihle preserving a high value of BAC.

For unlearning effectiveness, all approximate methods achieve close values relative to Retrain, with differences in TBAC below 0.02 across all scenarios. The MIA metric shows that SalUn-CRA maintains values closest to Retrain, indicating unlearning behavior closer to the gold standard under privacy metrics.


\subsection{Clinical Risk Analysis}

\begin{figure}[h]
    \centering
    \includegraphics[width=0.7\linewidth]{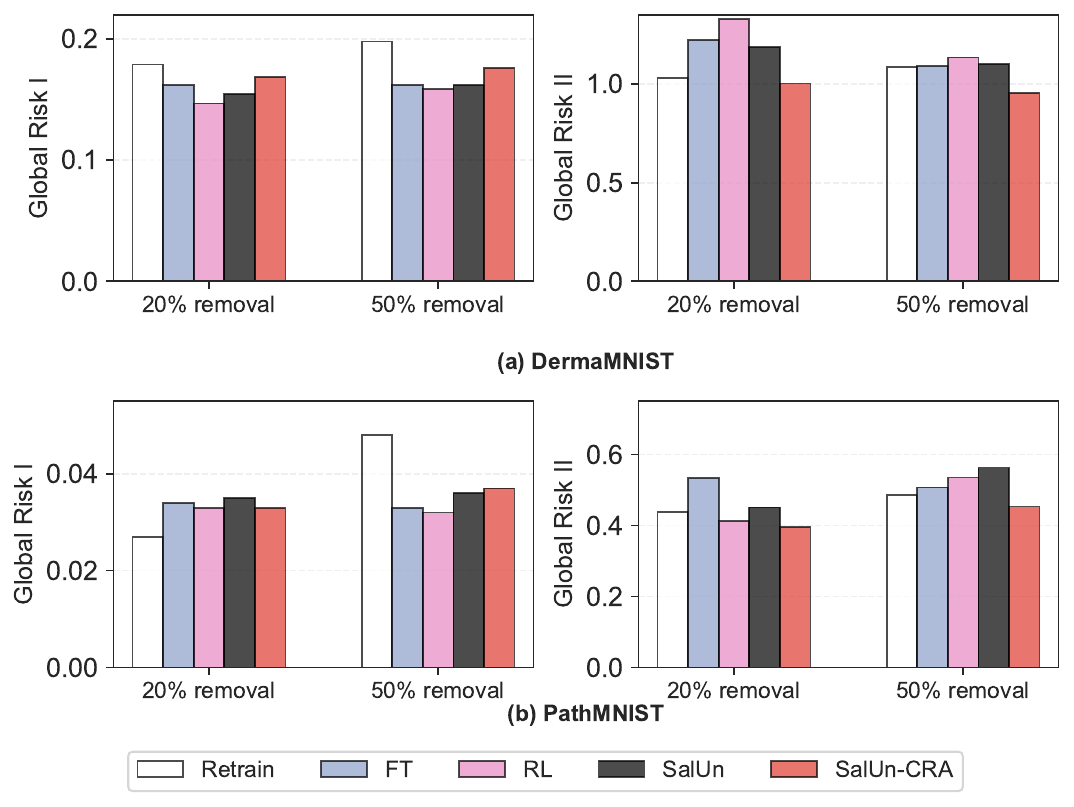}
    \caption{Clinical risk comparison across unlearning methods with forget rates 20\% and 50\% on DermaMNIST and PathMNIST datasets. }
    \label{fig:barras_risco}
\end{figure}

Figure~\ref{fig:barras_risco} presents clinical risk metrics across all experimental scenarios. 

On DermaMNIST, FT, RL and SalUn tend to reduce Global Risk I compared to Retrain, but at the cost of increasing Global Risk II. This pattern reveals a concerning trade-off: these methods favour specificity over recall, leading to more missed malignant lesions. For instance, at 50\% removal, RL achieves the lowest Global Risk I but exhibits elevated Global Risk II due to reduced sensitivity.
SalUn-CRA achieves Global Risk II values lower than the standard Salun in both removal scenarios, but higher results when considering Risk I. 

On PathMNIST, similar patterns emerge with smaller absolute differences. SalUn-CRA consistently achieves the lowest Global Risk II among all methods, including Retrain. 

\subsection{Trade-off Between Unlearning Effectiveness and Clinical Risk}

\begin{figure}[t]
    \centering
    \includegraphics[width=0.7\linewidth]{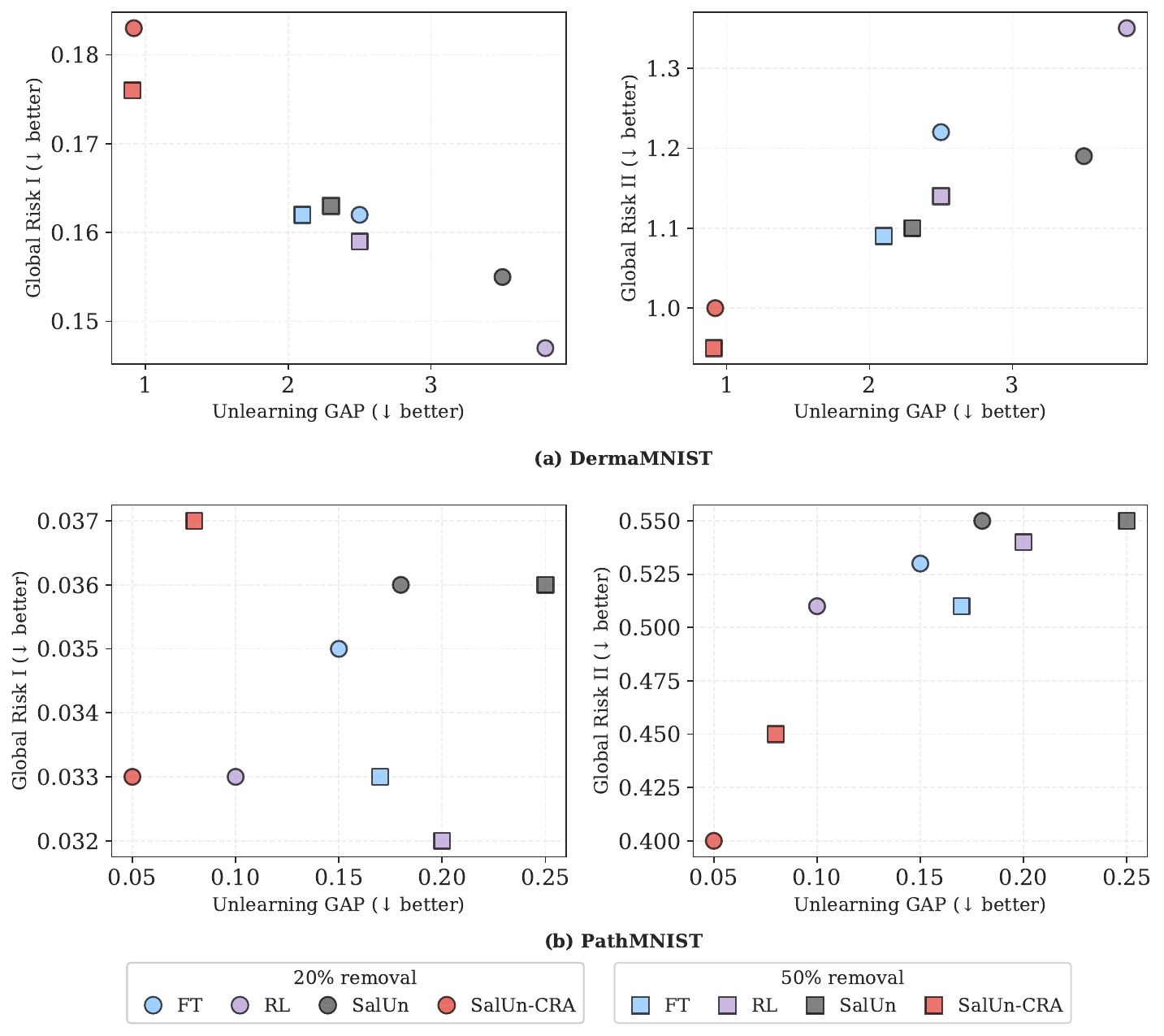}
    \caption{Trade-off between unlearning effectiveness (GAP) and Global Risk. Lower-left region represents ideal performance (low GAP and low risk). Circles indicate 20\% removal; squares indicate 50\% removal. }
    \label{fig:scatter_risco}
\end{figure}

Figure~\ref{fig:scatter_risco} illustrates the relationship between unlearning effectiveness (measured by GAP) and clinical risk. The ideal operating point is located in the bottom-left region of each plot, representing simultaneously low GAP (close approximation to Retrain behavior) and low Global Risk.

On DermaMNIST, SalUn-CRA (red markers) consistently positions itself in the favorable region for Global Risk II, achieving both low GAP and low risk. In contrast, methods such as RL achieve lower Global Risk I but exhibit higher GAP values and substantially higher Global Risk II, indicating that their apparent risk reduction comes at the cost of divergent model behavior and increased false negatives.

For PathMNIST, SalUn-CRA maintains its position as the method with the best trade-off for the most risky scenario, achieving the lowest Global Risk II while preserving near-zero GAP across both removal scenarios. These results demonstrate that SalUn-CRA successfully navigates the tension between unlearning effectiveness and clinical safety, occupying the Pareto-optimal region where other methods fail to reach.

\section{Discussion}

Our results highlight an important limitation in current machine unlearning validation protocols. While approximate unlearning methods such as SalUn and Random Labeling may maintain competitive performance under standard metrics (TBAC, MIA), they can increase clinical risk under asymmetric cost settings due to elevated false negative rates.


The central finding of this work is that standard machine unlearning methods can amplify clinical risk by disproportionately affecting sensitivity to malignant cases. When a data removal request is processed using conventional methods such as Fine-Tuning or Random Labeling, the model tends to become more conservative, favoring specificity over recall. While this may appear beneficial in terms of overall accuracy or balanced metrics, it translates to more missed diagnoses in practice.

This finding has direct implications for healthcare systems implementing \textit{the right to be forgotten}. A naive deployment of unlearning algorithms could inadvertently compromise patient safety, creating a tension between privacy compliance and diagnostic reliability that must be carefully managed.

\section{Conclusion}
\label{sec:conclusion}

In this work, we investigated the impact of machine unlearning on clinical risk in medical image classification. 
We introduced Global Risk I and II as primary evaluation criteria, explicitly modeling the asymmetric cost structure of medical diagnosis. We proposed SalUn-CRA (Clinical Risk-Aware), a modification of SalUn that replaces random relabeling with entropy-based forgetting for malignant samples. Experiments on DermaMNIST and PathMNIST under 20\% and 50\% balanced removal demonstrate that SalUn-CRA reduces clinical risk while maintaining competitive unlearning performance.

Our findings emphasize that clinical risk should be considered a first-class evaluation criterion in medical unlearning research. Incorporating cost-sensitive validation into unlearning pipelines is essential to ensure that regulatory compliance does not compromise patient safety.


\bibliographystyle{sbc}
\bibliography{sbc-template}

@article{warnecke2021,
  title={Machine unlearning of features and labels},
  author={Warnecke, Alexander and others},
  journal={arXiv preprint arXiv:2108.11577},
  year={2021}
}

@incollection{Chan2020,
  author    = {Chan, Heang-Ping and Samala, Ravi K. and Hadjiiski, Lubomir M. and Zhou, Chuan},
  title     = {Deep Learning in Medical Image Analysis},
  booktitle = {Deep Learning in Medical Image Analysis: Challenges and Applications},
  pages     = {3--21},
  publisher = {Springer},
  year      = {2020}
}

@misc{gdpr,
  author       = {{European Parliament and Council of the European Union}},
  title        = {Regulation (EU) 2016/679 (General Data Protection Regulation -- GDPR)},
  howpublished = {\url{https://eur-lex.europa.eu/eli/reg/2016/679/oj}},
  year         = {2016},
  note         = {Accessed: 25 Feb. 2026}
}

@misc{lgpd,
  author       = {{Brazil}},
  title        = {Brazilian General Data Protection Law (Law No. 13,709/2018)},
  howpublished = {\url{https://www.planalto.gov.br/ccivil_03/_ato2015-2018/2018/lei/l13709.htm}},
  year         = {2018},
  note         = {Accessed: 25 Feb. 2026}
}

@article{Hoofnagle2019,
  author  = {Hoofnagle, Chris Jay and van der Sloot, Bart and Borgesius, Frederik Zuiderveen},
  title   = {The European Union General Data Protection Regulation: What It Is and What It Means},
  journal = {Information \& Communications Technology Law},
  volume  = {28},
  number  = {1},
  pages   = {65--98},
  year    = {2019},
  doi     = {10.1080/13600834.2019.1573501}
}

@inproceedings{cordeiro25,
 author = {Andreza Falcao and Filipe Cordeiro},
 title = { Análise de Desaprendizado de Máquina em Modelos de Classificação de Imagens Médicas},
 booktitle = {Anais Estendidos do XXV Simpósio Brasileiro de Computação Aplicada à Saúde},
 location = {Porto Alegre/RS},
 year = {2025},
 issn = {2763-8987},
 pages = {43--48},
 publisher = {SBC},
 address = {Porto Alegre, RS, Brasil},
 doi = {10.5753/sbcas_estendido.2025.6966},
 url = {https://sol.sbc.org.br/index.php/sbcas_estendido/article/view/35582}
}

@article{barez2025open,
  title={Open Problems in Machine Unlearning for AI Safety},
  author={Barez, Fazl and Fu, Tingchen and Prabhu, Ameya and Casper, Stephen and Sanyal, Amartya and Bibi, Adel and O'Gara, Aidan and Kirk, Robert and Bucknall, Ben and Fist, Tim and Ong, Luke and Torr, Philip and Lam, Kwok-Yan and Trager, Robert and Krueger, David and Mindermann, S{\"o}ren and Hernandez-Orallo, Jos{\'e} and Geva, Mor and Gal, Yarin},
  journal={arXiv preprint arXiv:2501.04952},
  year={2025}
}

@article{zhang2023review,
  title={A review on machine unlearning},
  author={Zhang, Haibo and Nakamura, Toru and Isohara, Takamasa and Sakurai, Kouichi},
  journal={SN Computer Science},
  volume={4},
  number={4},
  pages={337},
  year={2023},
  publisher={Springer}
}

@article{chan2020deep,
  title={Deep learning in medical image analysis},
  author={Chan, Heang-Ping and Samala, Ravi K and Hadjiiski, Lubomir M and Zhou, Chuan},
  journal={Deep learning in medical image analysis: challenges and applications},
  pages={3--21},
  year={2020},
  publisher={Springer}
}

@misc{medmnist,
  author    = {Yang, Jiancheng and Shi, Rui and Wei, Donglai and Liu, Zeju and Wang, Li and Zhou, Yifan and Zhou, Sheng and Bian, Chao and Li, Lei and Wang, Xiaodan and others},
  title     = {MedMNIST: A Lightweight AutoML Benchmark for Medical Image Analysis},
  year      = {2021},
  howpublished = {\url{https://medmnist.com}},
  note      = {Accessed: February 13, 2025}
}

@article{li2024taxonomy,
  author    = {Na Li and Chunyi Zhou and Yansong Gao and Hui Chen and Anmin Fu and Zhi Zhang and Shui Yu},
  title     = {Machine Unlearning: Taxonomy, Metrics, Applications, Challenges, and Prospects},
  journal   = {ACM Computing Surveys},
  year      = {2024},
  url       = {https://doi.org/10.1145/nnnnnnn.nnnnnnn}
}

@inproceedings{hardan2025forget,
  title={Forget-MI: Machine Unlearning for Forgetting Multimodal Information in Healthcare Settings},
  author={Hardan, Shahad and Taratynova, Darya and Essofi, Abdelmajid and Nandakumar, Karthik and Yaqub, Mohammad},
  booktitle={International Conference on Medical Image Computing and Computer-Assisted Intervention},
  pages={204--213},
  year={2025},
  organization={Springer}
}

@inproceedings{bourtoule2021machine,
  title={Machine unlearning},
  author={Bourtoule, Lucas and Chandrasekaran, Varun and Choquette-Choo, Christopher A and Jia, Hengrui and Travers, Adelin and Zhang, Baiwu and Lie, David and Papernot, Nicolas},
  booktitle={2021 IEEE symposium on security and privacy (SP)},
  pages={141--159},
  year={2021},
  organization={IEEE}
}

@inproceedings{maverick2025,
  title     = {Maverick: Collaboration-Free Federated Unlearning 
               for Medical Privacy},
  author    = {Deng, Zhipeng and others},
  booktitle = {Lecture Notes in Computer Science},
  publisher = {Springer},
  year      = {2025}
}

@inproceedings{sakib2024unlearning,
  title     = {Machine Unlearning in Digital Healthcare: Addressing 
               Technical and Ethical Challenges},
  author    = {Sakib, Shaik Khairul and Xie, Mingqian},
  booktitle = {Proceedings of the AAAI Symposium Series},
  volume    = {4},
  number    = {1},
  pages     = {319--322},
  year      = {2024},
  doi       = {10.1609/aaaiss.v4i1.31809}
}

@inbook{ling2010cost,
  title     = {Cost-Sensitive Learning and the Class Imbalance Problem},
  author    = {Ling, Charles X. and Sheng, Victor S.},
  booktitle = {Encyclopedia of Machine Learning},
  editor    = {Sammut, Claude and Webb, Geoffrey I.},
  pages     = {231--235},
  year      = {2010},
  publisher = {Springer}
}

@inproceedings{fan2024salun,
  author    = {Chongyu Fan and Jiancheng Liu and Yihua Zhang and Eric Wong and Dennis Wei and Sijia Liu},
  title     = {SalUn: Empowering Machine Unlearning via Gradient-Based Weight Saliency in Both Image Classification and Generation},
  booktitle = {International Conference on Learning Representations (ICLR)},
  year      = {2024},
  url       = {https://arxiv.org/abs/2310.12508}
}

@article{nasirigerdeh2024machine,
  title={Machine Unlearning for Medical Imaging},
  author={Nasirigerdeh, Reza and Razmi, Nader and Schnabel, Julia A and Rueckert, Daniel and Kaissis, Georgios},
  journal={arXiv preprint arXiv:2407.07539},
  year={2024},
  url={https://arxiv.org/abs/2407.07539},
  note={Acesso em: 23 fev. 2025}
}

@inproceedings{golatkar2020eternal,
  title={Eternal sunshine of the spotless net: Selective forgetting in deep networks},
  author={Golatkar, Aditya and Achille, Alessandro and Soatto, Stefano},
  booktitle={Proceedings of the IEEE/CVF conference on computer vision and pattern recognition},
  pages={9304--9312},
  year={2020}
}

@inproceedings{graves2021amnesiac,
  title={Amnesiac machine learning},
  author={Graves, Laura and Nagisetty, Vineel and Ganesh, Vijay},
  booktitle={Proceedings of the AAAI Conference on Artificial Intelligence},
  volume={35},
  number={13},
  pages={11516--11524},
  year={2021}
}

@article{wu2019wider,
  title={Wider or deeper: Revisiting the resnet model for visual recognition},
  author={Wu, Zifeng and Shen, Chunhua and Van Den Hengel, Anton},
  journal={Pattern Recognition},
  volume={90},
  pages={119--133},
  year={2019},
  publisher={Elsevier}
}

@article{mester2024medical,
  title={Machine Unlearning for Medical Imaging},
  author={Mester, Soufiane and et al.},
  journal={ResearchGate},
  year={2024},
  note={}
}

@article{haimerl2025,
   author    = {Haimerl, Martin and Reich, Christoph},
   title     = {Risk-based evaluation of machine learning-based 
                classification methods used for medical devices},
   journal   = {BMC Medical Informatics and Decision Making},
   volume    = {25},
   number    = {1},
   pages     = {126},
   year      = {2025},
   doi       = {10.1186/s12911-025-02909-9}
 }

@inproceedings{scholz2024imbalance,
  title={Imbalance-aware loss functions improve medical image classification},
  author={Scholz, R. and et al.},
  booktitle={Proceedings of Machine Learning Research},
  year={2024},
  note={}
}

\end{document}